Erkki Luuk[†]

# A type-theoretical approach to Universal Grammar

**Abstract.** The idea of Universal Grammar (UG) as the hypothetical linguistic structure shared by all human languages harkens back at least to the 13th century. The best known modern elaborations of the idea are due to Chomsky. Following a devastating critique from theoretical, typological and field linguistics, these elaborations, the idea of UG itself and the more general idea of language universals stand untenable and are largely abandoned. The proposal tackles the hypothetical contents of UG using dependent and polymorphic type theory in a framework very different from the Chomskyan ones. We introduce a type logic for a precise, universal and parsimonious representation of natural language morphosyntax and compositional semantics. The logic handles grammatical ambiguity (with polymorphic types), selectional restrictions and diverse kinds of anaphora (with dependent types), and features a partly universal set of morphosyntactic types (by the Curry-Howard isomorphism).

## 1. Introduction

Although the idea of Universal Grammar (UG) harkens back to at least Roger Bacon (cf. Ranta 2006), the modern version of the hypothesis is usually credited to Chomsky (e.g. 1970, 1981, 1995). In modern times, the notion of UG has taken several forms: a substantive, diluted and implicational one (Chomsky 1981, Jackendoff 2002, Greenberg 1966, resp.). However, a logical (derivational) path from implicational universals (structural traits implying other structural traits) to functional dependencies to substantive universals has been overlooked. The present paper tries to unveil this possibility in the form of a direct type-theoretical account of substantive universals as logical formulas.

From the present viewpoint, type theory is essential for a logical specification of UG. First, in its modern form (i.e. as dependent and/or polymorphic type theory), it is the most expressive logical system (contrasted with the nonlogical ones such as set and category theory). In a logical approach (i.e. in one with simpler alternatives such as ZOL, FOL, SOL and HOL), complex type theories outshine simpler ones in accounting for phenomena like anaphora, selectional restrictions, etc. (Asher 2014; Luo 2010, 2014; Ranta 1994). Second, as the notion of type is inherently semantic:

(1) type ≔ category of semantic value,

it is by definition suited for analyzing universal phenomena in natural language (NL), as NL semantics in largely universal (as witnessed by the possibility of translation from any human language to another). Thus, if one could build a fundamentally semantic description of grammar (e.g. one on top of and integrated with a semantically universal description of NL), it would at least stand a chance of being universal.

The paper will proceed as follows. After a brief overview, we introduce a simple logic (say, L) matching NL morphosyntax as closely as possible. The purpose of L is to express the universal core of NL morphosyntax in a formally and linguistically precise, yet parsimonious way. In the second part, L is furnished with elementary and complex morphosyntactic types, some of which are universal, others nonuniversal or possibly universal. The third part gives additional explanations, examples and rules, situating L within dependent and polymorphic type theory along the way (with Martin-Löf's type theory (MLTT – Martin-Löf 1984) as the main type-theoretical reference point).

---

† Institute of Computer Science, University of Tartu, Estonia. Email: erkkil@gmail.com

## 2. Preliminaries

Returning to modern notions of UG, there are, then, three to consider (some of which are not mutually exclusive): one of an innate language acquisition device (LAD), the second of implicational universals, and the third of the hypothetical linguistic structure shared by all human languages. The details of implicational universals imply a (larger than singleton) set of universal grammars, so the second notion is irrelevant if one insists (as we do) on a single UG. Thus we have two alternatives, the LAD which can be termed "weak" and the substantive universals that amount to a "strong" UG (an even stronger version can be constructed by the condition "the linguistic structure shared by all possible human languages"; however, we only indicate this possibility here and speak of a "strong UG" in the more lax sense henceforth). While the very fact that normal human infants acquire NL without an explicit supervision (seen as disjoint from normal exposition to linguistic stimuli) is an evidence for a LAD, an evidence for substantive universals is much more contentious. More recently, the strong version of the hypothesis, having suffered heavy blows from the sides of both linguistic theory (e.g. Jackendoff 2002) and comparative typological evidence (e.g. Evans and Levinson 2009), has been severely discredited and at least partly demolished and abandoned. In this paper, we use dependent and polymorphic type theory to set up a credible case for a strong UG, resulting in a framework very different from Chomskyan ones. In a sense, our approach will be more formal; secondly, the usual (although frequently implicit (and perhaps even inessential)) Chomskyan notion of syntax-as-grammar will be supplanted by morphosyntax-as-grammar, where, moreover, "morphosyntax" will be fundamentally "semantical" in nature. But let us start by introducing some key concepts.

In a nutshell, the picture is as follows: mathematically speaking, there are relations with arities and associated arguments; some relations, their arguments and the resulting formulas correspond to morphosyntactic constituents. By a morphosyntactic constituent we mean a well-formed (abbreviated wf) formula (abbr. wff) of a NL expression. For example, in

(2) THE(man),
(3) RUN(THE(Y(man))),
(4) (Y(LOVE))(mary,john),
(5) MAN(the)
(6) M(the)AN

(2)-(4) are wf formulas of *the man*, *the men run* and *john loves mary*, resp., while (5)-(6) are ill-formed ((6) is already a notational gibberish). (2)-(4) make use of the following conventions:

(7) Complex formulas are written in prefix notation, *A*(*B*) or (*A*)(*B*), with *A* standing for a relation and *B* for its argument(s),
(8) $0^{th}$ order relations ($1^{st}$ order arguments) in small letters, all other relations in capitals,
(9) Inside out and right to left valuation (or derivation); inside out has precedence over right to left,
(10) Argument derivation order SOD (subject, object, ditransitive object), otherwise linear.

Y is a tense/person/number (etc.) marker; we will make a precise sense of this term later. By (8), we see that *the man* is expressed as a first-order formula. To keep the representation in close correspondence with NL, we avoid extralinguistic and theory-specific features, such as the model-theoretic variable *x* in THE(MAN(*x*)). In many cases, as in this one, such features can be added later if a specific extralinguistic interpretation is desired. By (7), (9), we get *john loves mary* rather than vice versa for (4), and the person-number relation marker *-s* will apply to LOVE rather than to LOVE(mary,john) (the ungrammatical or incomplete English expression *john love mary*). By (10), arguments are derived in the order (subject, object, ditransitive object) if they have these values and in their linear order in the NL expression if not; by (9) this will specify tuples (D,O,S) or (...,3rd,2nd,1st), resp. (cf. (17)).

Our vocabulary consists of elementary relation symbols, commas and (matching) parentheses. There are now two (sensible) possibilities:

(11) For a particular language, the symbols are type constants; cross-linguistically they are type variables (e.g. 'man' valuates to *man* in English and *homme* in French).
(12) The symbols are type constants inhabited by language-particular (proof) objects (e.g. the type 'man' has objects *man*, *homme,* etc.).

Notice that (12) is more universal not only linguistically but already type-theoretically (by identifying formulae with types and proofs (more precisely, proof objects) with terms or objects, by the Curry-Howard isomorphism)[1]. Elementary relation symbols can be composed into well-formed relation formulas (henceforth usually just "formulas" or "wffs"); we also assume

(13) All (elementary) relation symbols are atomic formulas and all formulas signify relations,
(14) A wff is wf both syntactically and semantically, i.e. wf and well-typed (cf. section 7).

Formulas are morphosyntactic types that have semantics (by (1)). Our maximal (and default) universe of morphosyntactic types will be denoted **U**; in **U**, all formulas are morphosyntactic types and all types are morphosyntactic formulas. Later we will also consider other universes, e.g. **Rel** and **Arg** (types of morphosyntactic relations and arguments, resp.).

### 3. Discussion

This seemingly rather rudimentary representation bears some remarkable features. First, it can capture certain aspects of syntax, morphology as well as semantics; moreover, it manages to do so without a cumbersome graphical representation (e.g. trees). All words and morphemes (even those fused with stems like the plural marker in *men*) have meaning (i.e. semantics); hence the representation is compositional in both morphosyntactic and semantic respects (the exact nature of the compositionality will be made more precise later). For example, Y is a morphological (although perhaps also a syntactic) category in English. Obviously, one could make it more precise by substituting Y with 3SG (third person singular) in (4); the main reason for us not doing so is a tradeoff between such precision and universality.

For most languages, the representation partially captures even information structure and word order. The general principles

(15) Argument terms must be derived before the relation terms they are arguments of,
(16) Arguments depend on relations but the 'proof of a relation' (i.e. 'relation term') depends on 'proofs of its arguments' ('argument terms'), have ramifications for both information structure and word order. Principles (7), (9), (15) imply that arguments tend to be on the right and relations on the left; this order is reversed in the linear order of writing and speech ("preceding" and "succeeding" in writing and speech correspond to "on the right" and "...left" in the notation, resp.).

---

[1] It is also more universal in a general sense, as function is a subcase of relation (functions are certain sets of pairs; relations are all sets of *n*-tuples for any constant *n*).

The fact that L may be relational as well as functional (depending on the choice between (11)-(12), capital-letter formulas may stand for relation or function symbols) is noteworthy as well. Besides being largely interdefinable, both variants have their own merits (broadly speaking, generality for relational and specificity for functional logic – see Oppenheimer and Zalta 2011); in addition, if one prefers a functional interpretation, (s)he must choose between different potentially relevant function types (notably, between truth-functions and those valuating to linguistic expressions). For example, if one interprets (4) as truth-functional, any further morphosyntactic compositionality (as evident in e.g. *john loves mary and vice versa*) would be barred, as one would be left with only a truth value after valuating *john loves mary* (compositional **truth-functional** semantics would still be attainable with standard logical operators OR, NOT, etc., though). Since the arguments of truth-functions may be expression-functions but generally not vice versa, a truth-functional interpretation can be composed with an expression-functional one (but generally not vice versa), so it is desirable to refrain from truth-functionality for generality of one's logic.

## 4. Specifying formulas, deriving expressions

Let us agree on some terminology:

(17) 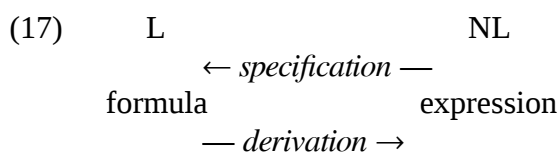

L        NL
    ← *specification* —
formula        expression
    — *derivation* →

We specify L's formulas from NL expressions and derive NL expressions from L's formulas. The derivation is not a function to NL as such; however, it should be a function to particular NLs (the precise character of this function remains open). An obvious goal is to make specification a function[2]; if successful, its possible injectivity would be another open question. Although we do not have any functions (yet), we will demonstrate an algorithm. This is neither a specification nor a derivation algorithm (which we presently cannot have), only one deciding a derivation's correctness. However, we can say something about the desired derivation algorithm as well. For the algorithm to consistently derive all and only the relevant well-typed expressions, it should adhere to the principles of

(18) Serial and incremental derivation of sentences (deriving one expression per stage using all previously derived expressions),
(19) Deriving a well-typed result at all stages of derivation,
(20) Inhabiting as many complex subformulas of the endformula as possible,
(21) (18) having precedence over (9).

Intuitively, (15), (18)-(19) could be viewed as following from Frege's principle of compositionality (this aside, they are stipulated for simplicity and wellordering – see section 11.3). (20) is a metric of the quality of the algorithm (by (18), some complex expressions may have to be assumed rather than derived (i.e. require a separate derivation)). By a 'stage of derivation' we mean (by (11)) the value or (by (12)) the proof object (or term) of a type, i.e. a linguistic expression. For example, the derivation from THE(man) is 2-staged, with the first deriving *man* and second *the man*; we will write the derivation "man > the man" or "m > tm" in the abbreviated form.

Consider now the following hypothetical logical forms of *i know the man who was ill*:

---
2   Since L's formulas depend on NL, they cannot be specified mechanically with an inductive definition; similarly, derivation is incomplete due to e.g. language-particular word order constraints.

(22) KNOW(WHO((Y(BE))(ill,THE(man))),i)
(23) KNOW((WHO((Y(BE))(ILL))(THE(man))),i)
(24) KNOW((WHO(ILL)(Y(BE)))(THE(man))),i)
(25) KNOW(WHO(Y((BE)(ILL))(THE(man))),i)
(26) KNOW(WHO(Y(BE)(ILL)(THE(man))),i)
(27) Y(ILL(BE(who)))(KNOW(THE(man),i))
(28) Y(ILL(BE(who)))(KNOW((THE(man),i))
(29) (Y(ILL(BE(who))))(KNOW((THE(man),i))
(30) KNOW(THE(man),i)(ILL(Y(BE(who)))
(31) KNOW(THE(man),i)(Y(ILL(BE(who))))
(32) KNOW(THE(man),i)Y(ILL(BE(who)))

Since (32) is not a formula by (7), it is immediately disqualified (the subformula Y(ILL(BE(who))) is not in brackets, i.e. not composed with KNOW(THE(man),i)). Likewise, (30)-(31) are not formulas (by (9), (THE(man),i) must be applied to the relation (…) on its right before applying KNOW to (THE(man),i), but the latter, a 2-tuple of arguments, is not a formula, so (30)-(31) are not formulas. (26) is disqualified by (19) (by deriving the ill-typed *the man ill* on the 3[rd] stage); similarly, (25) derives the ill-typed *the man be ill* on the 3[rd] stage. (22)-(23) yield the same derivation: m > tm > tmwi > tmwwi > iktmwwi. (24) yields m > tm > tmw > tmwi > tmwwi > iktmwwi, one more than (22)-(23). (27) yields i > iktm > iktmwwi, disqualified by (20); (28) yields m > tm > iktm > iktmw<u>b</u>i, the ill-typed term is underlined; (29) derives m > tm > iktm > iktmwwi, disqualified by (20). The winner (by (20)) is (24). However, since NL grammar does not have to follow (20), it is unclear which, if any, of (ILL)(Y(BE)), (Y(BE))(ILL) and (Y(BE))(ill,THE(man)) (resp., (24), (23) or (22)) is grammatical. We will return to this problem in section 8.

## 5. Generalized anaphora

I propose the term "generalized anaphora" for analyzing PRO, C (see Tab. 2) and agreement in context. Consider

*he, me, i, they, it...* : PRO,
*that, which, who, what, when, where, why...* : C,

where, as usual, the typing judgement "is/are of type" is abbreviated ":". In order to refer, PROs and Cs must corefer; some examples of coreference are given below (we allow *ad hoc* abbreviations of relation symbols to fit formulas on a single line):

– *every farmer who owns a donkey beats it* : $(Y_p(BEAT))(it_y,((Y_x(OWN))(A(don_y),who_p))(EV(fa_p)))$.
(33) *every farmer beats a donkey he owns* : $(Y_f(OWN))(he_f,(Y_f(BEAT))(A(don_y),EV(fa_f)))$.
– *i know the man who was ill* : $KNOW((WHO_a(ILL)(Y_a(BE)))(THE(man_a)),i)$.
(34) *it rains* : $(Y_d(RAIN))(it_d)$.

Coreference (or -indexing) is done with small subscript letters. This is an optional feature of the representation (which we will occasionally suppress for simplicity); however, it is necessary for capturing agreement and for tracking in rules (50), (52). In (34), *it* is a dummy pronoun, required by English but not e.g. Estonian or Spanish syntax, where the formula would be (Y(RAIN)) (RAIN in capitals because it is a zeroary relation symbol in these languages). In every language, a relation (e.g. Y, A or OWN) has a fixed arity (number of arguments; in this case, one, one, two, resp.); in rare cases the arity of a relation (e.g. RAIN) in two languages may differ (cf. Evans and Levinson 2009:431). (In English, *rain* is also a noun; such polymorphisms are analyzed in section 11.4.) In sum, any relation has a fixed arity per language; this is a precondition (not only) for formulas' well-formedness.

For (33), one might be tempted to consider the alternative formula

(35) $(Y_z(OWN))((Y_z(BEAT))(A(don_y),he_z),EV(fa_z))$.

Since (33) derives f > ef > efbod > efbodho but (35) f > ef > efbod̲h̲ > efbodho, the correct one (by (19)) is (33) (the word making the third derivation stage ill-typed is underlined).

## 6. Quantifiers

Since quantifiers are higher-order relations (Westerstahl 2011), i.e. an $n^{th}$ order quantifier is an $n+1^{th}$ order relation, they are straightforwardly implemented as such in the formulas (assuming their presence in the expression). By 'quantifiers' I mean those in the usual linguistic sense of the word, i.e. expressions like *one, two, million, much, many, all, every, few, no, some*, etc., so we will have considerably less quantifiers than in the theory of generalized quantifiers, where even a proper name like *john* would be a (type <1>) quantifier (Keenan and Westerstahl 2011). Here are some examples involving quanfiers (see also (33)):

*a million men went to work* : $(Y(G))(TO(work),A(MIL)(Y(m)))$,
*at least five but less than eight men left* : $(Y(LEAV))(BUT(LESS\_TH(EI),AT\_LE(FI))(Y(m)))$.

More complex instances of quantification will be considered in section 9.3.

## 7. Selectional restrictions

The next (optional) step is to add selectional restrictions to our formalism (cf. Asher 2014, Luo 2010). It is reasonable to suspect that an A like *red* imposes the selectional restriction that its argument be of type physical entity (P) or that a V like *read* imposes a restriction that its subject be a sentient and object an informational entity (i.e. of type S and I, resp.). This will be expressed by

$RED_P(x)$ and $READ_{I,S}(x,y)$,

with capital subscript letters imposing type restrictions on the relation's argument(s); $READ_{I,S}(x,y)$ means that *x* is restricted to type I and *y* to type S. Such restrictions are followed by default but not always; e.g. one may say *red ideas* in a metaphorical sense of communist ideas, etc.; for such contexts we stipulate the rule (with all letters denoting constants):

(36) $$\frac{X,\ y,\ t,\ Z_s(y)\ \text{type} \qquad \neg\text{wf}(X_t(y)) \qquad \exists Z \exists X(\llbracket Z \rrbracket \mapsto X)}{\llbracket Z_s(y) \rrbracket \leq: \llbracket X_t(y) \rrbracket\ \text{type}}$$

where $A_b(c)$ is a morphosyntactic relation type (with selected argument type indicated in subscript), $\llbracket A_b(c) \rrbracket$ the corresponding semantic interpretation, wf the predicate "is well-formed" and ≤: the relation "is a subtype of". For example, differently from *communist ideas*, *red ideas* is morphosyntactically ill-typed but by virtue of (36) and the function $\llbracket COMMUNIST \rrbracket \mapsto RED$, any interpretation of *communist ideas* is a subtype of the (totality of) interpretation of *red ideas*. One needs (something like) rule (36) to explain metaphor and metonymy without tampering with selectional restrictions. Below are some example formulas with selectional restrictions (A ≔ animate entity; note that the second argument of *read* is optional):

*you are ill* : $(Y_a((BE)(ILL_A)))(you_a)$,
*john reads* : $(Y(READ_{I,S}))(john)$.

## 8. Rules

It is now time to formulate some rules by which morphosyntactic types compose. Quite inevitably, all languages have morphosyntactic relations (and their arguments); the question is whether we can assign them universal types in any useful way. We propose the following:

*Tab. 1. Rules*

*john, john smith, ontario, lake ontario...* : **PN**
*apple, mole, cat, tree...* : N
*(s)he, me, i, they, it...* : PRO
*that, which, who(m), what, when, where, why...* : C
*round, love, sleep, drink, kill, walk, taste, smell, run...* : X/R/*
*be, seem, become, taste, smell, appear...* : COP
*start, attempt, yearn...* : IR
*have, do, be, can, may...* : AUX
*agree, detest, justify, eat, be, seem, become, appear, conceive, summon, consider...* : V
o(V, COP, IR, AUX, R/*) : **R**
(37) o(N, PN, PRO, X/*, GER) : **X**
*this, that, those, these...* : DEM
*a, the, other...* : DET
(38) o(DET, DEM) : **D**
*one, two, zero, sixty, million...* : NUM
o(NUM), *all, some, many, every, a pile of, more (than), less (than), (at) least, most (of)...* : **Q**
o(tense, aspect, mood, voice) : **Yr**
(39) o(person, gender, number) : Yx
(40) ⟨Y(x|r)⟩ = **Y**
*beautiful, wise, clumsy, neat, horrible...* : A
o(R-*ing*...) : GER/PCJ/PCV
(41) o(INF1, …, INF*n*) : INF
o(NOM, ACC, GEN, LOC...) : CAS
o(GEN), *of...* : **POS**
*at, to, with, without, by, behind...* : ADP
o(CAS, ADP) : **CA**
(42) *to him, she, john, his, with someone, on the picture, in a random place...* : [CA](CA(Xh)) = **CAP**
*very, distinctly, especially, somewhat...* : ADV
o(ADV, PCV), *at once, by chance, by and large, at the same time, in a hurry...* : **ADL**
(43) ⟨**ADL**⟩**([Y](R|A))**
*not, (n)either(...(n)or), and, but, or, if(...then), then...* : **CON**
(44) CON{.\{CON,Y,CAS}}* = **CONC**
(45) o(**{D,PS,⟨Q⟩,⟨CAP⟩,⟨[ADV|PCV](A|PCJ)⟩}*(⟨[Yx](X\PRO)⟩)**, PRO) : **XP**
(46) ⟨ADL⟩(Y(R)){CAP}* = **RS**

(47) {[Y](COP),CAP,⟨ADL⟩[(A|PCJ)],⟨Q⟩,PS}*(CAP) = **COPS**
(48) *(Y(IR)){⟨ADL⟩,INF*}* = IFS
(49) *(Y(AUX)){INF*,CAP}* = AUXS
(50) o($C_a(S\backslash Xh_a)$) : **IS**
(51) o(RS, IS, IFS, AUXS, COPS) : **S**
(52) *that this is so, the man who smiled...* : $C_a(S[\backslash Xh_a])$ = **XC**
(53) o(XP, XC) : **Xh**
*however, moreover, since, because...* : S-CON
S-CON(S,S)

*Tab. 2. Definitions*

$B \sqsubseteq A$ ≔ $A$ subsumes $B$.
$A|B$ ≔ an element from $\{A,B\}$.
$[A]$ ≔ $A$ is optional.
$[p]\{A\}[q]$ ≔ operators $p, q$ over a set of formulas $A$, commas and parentheses ($A$ a comma-separated sequence of formulas).
$⟨A⟩$ ≔ an admissible (incl. empty) sequence of formulas $A$, commas and parentheses ($A$ a formula).
$A\backslash B$ ≔ $A$ except $B$.
$A/B$ ≔ the type polymorphic over types $A, B$.
. ≔ any formula in the universe (as specified in Tab. 1).
* ≔ any (incl. empty) string admissible in the context.
o($x$) ≔ all terms of type(s) $x$ ($x$ a sequence of types).
A ≔ adjective
ACC ≔ accusative
ADL ≔ adverbial
ADP ≔ adposition
ADV ≔ adverb
AUX ≔ auxiliary verb
AUXS ≔ auxiliary sentence
C ≔ complementizer
CA ≔ case|adposition
CAP ≔ case|adposition phrase
CAS ≔ case
CON ≔ connective
CONC ≔ connective composition
COP ≔ copula
COPS ≔ copular sentence
D ≔ determiner|demonstrative
DEM ≔ demonstrative
DET ≔ determiner
GEN ≔ genitive
GER ≔ gerund
IFS ≔ infinitival sentence
INF ≔ infinitive
IR ≔ infinitival relation
IS ≔ interrogative sentence
LOC ≔ locative
N ≔ noun
NOM ≔ nominative

NUM ≔ numeral
PCJ ≔ adjectival participle
PCV ≔ adverbial participle
PN ≔ proper name
POS ≔ possessive
PRO ≔ pronoun
PS ≔ possessive sequence (see section 9.3.3)
Q ≔ quantifier
R ≔ core relation
RS ≔ R-sentence
S ≔ sentence
V ≔ verb
X ≔ core argument (from XP)
XC ≔ argument clause
Xh ≔ higher-order argument
XP ≔ (noun|determiner) phrase
Y ≔ marker
Yr ≔ relation marker
Yx ≔ argument marker

### 9. Remarks and additional examples

The rules are preliminary and inexhaustive. First occurrences of claimed universal types are bold and possibly universal ones underlined. All estimates as to the (possible) universality are conservative. The rules, although partly (inter-)universal, have particular languages as their universes, so terms of different languages do not get mixed. o is a filter function accepting all linguistic expressions and returning those of the type of its specified argument(s) (the argument for "linguistic expression" is unspecified (suppressed in the notation)). (48), (49) are incomplete specifications. Gender and the grammatical category 'noun class' are synonyms. CON takes any one or two formulas (except CON, Y and CAS) as arguments. Yx marks properties of Xs; Yr marks properties of Rs. (40) posits a single type for all (incl. combined, e.g. Yx(Yr)) markers. (43) accounts for Yr marking on predicative As in languages like Korean, Lao, Semelai, etc. (Hajek 2004; Stassen 2008). We assume S-CONs to take two Ss as arguments even when they occur in the beginning of Ss (so in a S like *however, this does not compute*, "*however*" is taken to bind the previous sentence as well). Xh could be substituted with XC|XP but we have chosen to make the type explicit. Below are some additional example terms of types in Tab. 1 (more are given in the following sections):

– *john, stone, his son's more than 435345 mildly interesting books on syntax in paperback...* : XP
– *j sleeps; hurriedly, m sent her a letter; m sent her a short letter in a hurry; j loves m; someone killed him in a fit of rage; a surprisingly iridescent blue acorn was simultaneously seen by seven random rats in the garden behind the house; it rains; sajab; llueve...* : RS

### 9.1. Nonfinite R

The nonfinite R paradigm encompasses GER, PCJ, PCV and INFs; the first three are distinguished by their function (GER functions as "noun", PCJ as "adjective" and PCV as "adverb"; their English terms are only functionally distinct). Below are some examples:

– ***running*** *quickly, she finished first* (PCV)

– the quickly **running** water burrowed deep into the ground (PCJ)
– **running** is healthy (GER)
– i like **to run** (INF2)
– i must **run** (INF1)

It is another quirk of English that AUX and IR take different INFs (the bare and *to*-infinitive (typed by (41) INF1 and INF2, resp.)). We assume INFs to be arguments of finite Rs.

### 9.2. COPS

We need an account of Ss like

– j is (*william|nice|50|behind the house|her husband|walking*),

of which

– *j is walking* : (Y(WA))(j) <: RS,

is an RS rather than COPS (so we have to beware of this English-particular Y/PCJ polymorphism). For the rest there are some examples of (47):

(54) *i am in a hurry* : {Y(COP),ADL}*(CAP),
(55) *this was broken beyond repair*, *the man whom i saw was very ill* : {Y(COP),ADL(A)}*(CAP),
(56) *the bag is her farther's* : {Y(COP),PS}*(CAP),
(57) *he is 57* : {Y(COP),⟨Q⟩}*(CAP),
(58) *this is* (*him|william*) : {Y(COP),CAP}*(CAP),
(59) *eto chtenie* (*this* [is ]*reading*, Russian), *wait nglayayala* ([the ]*man* [is/was a ]*doctor*, Pitjantjatjara (Douglas 1959:55)) : {CAP}*(CAP).

Some COPs are Vs while other are R/*s. In *this was broken*..., Y encodes (among other grammatical categories) voice. The *(CAP) types in (54)-(59) are proper subtypes of COPS.

### 9.3. Self-composition

By self-compositions are meant single-type compositions (but not all of them). Consider

(60) *more than as many as eight thousand seven hundred fifty four*
*(61) at least as many as 89098 less than 989890 more*
(62) *his mother's parents' son's dog's...*
(63) *more than 8374874 of his mother's father's uncle's daughter's less than 7644874 estates' mansions' kitchens' at least 989898 stoves' chimneys' bricks'...*
(64) *she remembered the man who had said that the apples he had eaten were poisoned, something which had seemed strange at the time*
(65) *a strikingly iridescent small acorn*
(66) *an irrelevantly iridescent, small, blue, astonishingly symmetric acorn*.
*(67) a commander, thug, sailor, mercenary, fighter and captain*
(68) *he ran, jumped, rolled or crawled*
(69) *in a dull voice, a vague speech was given to the audience through loudspeakers in the woods*
*(70) i'm rather in a hurry*
*(71) he ran very quickly*

(60)-(71) are examples of nontrivial *n*-compositions of a single type in context ("nontrivial" means *n*>1). For self-compositions we additionally stipulate

(72) Self-compositions do not contain pausal CONs,
(73) By English ortography, pausal CONs are signaled by {., ,, ;, :, –}, the punctuation being a sufficient but not necessary condition for pausal CONs.

By (72)-(73), (66)-(68) are not self-compositions. Most self-compositions are denoted by ⟨...⟩ in Tab. 1; e.g. (70)-(71) are nontrivial examples of ⟨ADL⟩ ((71) is also an ⟨ADV⟩). In analyzing a self-composition, start by breaking it into single-type terms, e.g. Qs *more than, as many as, eight, thousand, seven, hundred, fifty, four*. Next, determine what modifies what, e.g. MT(AMA(8(1000)+7(100)+50+4)), where + is the implicit addition operator (encoded in NL as the concatenation of corresponding numerals). Skipping most details, we type the compositions ⟨Q⟩ or {⟨Q\NUM⟩,⟨NUM⟩}* (by Tab. 2, it is easy to see that ⟨Q⟩ = {⟨Q\NUM⟩,⟨NUM⟩}*).

### 9.3.1. CONC

There are several types of self-compositions in NL, with CONC and the clausal one most prominent. For example, an $n^{th}$-order self-composition with a binary CON for *n*>0 is CON(CON(*A,A*),CON(*B,B*)) (*A, B* types), usually (or always?) serialized as *A* CON *A* CON *B* CON *B* in NL (by (44), CON(CON(*A,A*),CON(*B,B*)) <: CONC). The terminology is as follows: 0-compositions are empty, 1-compositions are $0^{th}$ order, 2-compositions $1^{st}$ order, etc.

### 9.3.2. Clausal composition

The device for clausal composition is XC. From Tab. 1, we notice the hierarchy

(74) [Yx](X\PRO) ⊑ ⟨[Yx](X\PRO)⟩ ⊑ XP|XC ⊑ Xh ⊑ CAP ⊑ S ⊑ XC.

By Tables 1-2 and (13), each type to the right of a "⊑" in (74) is a relation over the type to the left of the "⊑". There is no circularity in (74), because the XC in XP|XC is of lower order than the rightmost one. For example, in the S *i know a man who thinks that a fish he caught was a frog*, the XC *who..* is defined through the S *a man thinks..*, etc., written (*A* {*B*} := *A* a relation over *B*):

– *who..* : XC {*a man thinks that..* : S {*a man, a fish he caught, a frog* : CAP {*a man, a frog* : Xh (XP), *a fish he caught* : Xh (XC) {*he caught a fish* : S {*he, a fish* : CAP {Xh (XP)}}}}}}.

Notice that the XC *a fish he caught* is of lower order than the XC *who thinks that a fish he caught was a frog*. To avoid complicating (52), we assume that XCs like *i know he left, a fish he caught*, etc., are elliptic abbreviations of *I know that he left, a fish that he caught,* etc.

### 9.3.3. PS

According to Heine (1997), POS is universal; if so, the same must apply to PS, suitably defined (as follows):

(75) PS = [(POS(PRO))]⟨[(POS([Y](X\PRO)))]⟩,

accounting for (62) and parts of (63). We assume that (at least in some contexts) the CA *of* is a POS

semantically equivalent ("≈") to genitive, so that e.g.

(76) "*j's x*" ≈ "*x of j*" : (POS(j))(x) <: XP.

Example (´ := GEN): *more than 8374874 of his mother's father's uncle's daughter's less than 7644874 boxes are empty* : ((Y(IS))(EM))(MT(8374874))(OF(((´(he))(´(M))(´(F))(´(U))(´(D))) (LT(7644874(Y(box)))))).

By (19), (76), the formula is *(´(he))(´(M))(´(F))(´(U))(´(D))* rather than *(´(D))(´(U))(´(F))(´(M)) (´(he))*. Notice that (19) requires the derived expression to be well-typed, not wf. Arguably, "*daughter's less than 7644874 boxes are empty*" is a well-typed construction without being a wf S. Omitting details on numerals, we obtain the following derivation: box > boxes > 7644874b > lt7644874b > dlt7644874b > udlt7644874b > fudlt7644874b > mfudlt7644874b > hmfudlt7644874b > ohmfudlt7644874b > 8374874ohmfudlt7644874b > mt8374874ohmfudlt7644874b > mt8374874ohmfudlt7644874bae.

## 10. Open problems

Below are some open problems, most of them having to do with representing NL in L (there are many more):

1. It is quite clear what are XP's constituents – CAP, Q, PS, D, [ADV|PCV](A|PCJ), [Yx](X\PRO); however, what modifies what is much more contentious (a similar argument applies for COPS).
2. Since As and PROs are claimed to be nonuniversal (Dryer 2013; Evans and Levinson 2009), we need a cross-linguistically universal characterization of A- and PRO-*like* words. For example, should we posit a supertype for A and R to do away with R|A in (43)?
3. Evans and Levinson (2009) claimed the nonuniversality of NUM. If so, how do languages without numerals account for numerosity – with approximating quantifiers like *few* and *many*? Does this mean that the languages do not support counting (and if they do, how)?
4. There should be more examples of cross-linguistic diversity to make the case about UG more credible. In particular, we need a comprehensive cross-linguistic analysis of the underlined types in Tab. 1. For example, some sources suggest C is missing in e.g. Chinese (see Yeung 2004, ch. 4).
5. What are *plus*, *minus*, *times*...? Quantifiers? If so, this would considerably complicate a comprehensive treatment of quantification.
6. (48), (49) are heavily underspecified (more examples of L's flexibility than useful rules) and need a lot more work.
7. The extent of the overlap between CAP and ADL\{ADV,PCV}. It would be useful (and make the system simpler) if we could identify the types. For example, in a S like *he walks behind the house*, is *behind the house* an argument of *walks* or vice versa? It could be either by (46).
8. The treatment of IS is incomplete even for English, and should (in all likelihood) account for the possibility of forming ISs with AUX. However, our representation is not well-suited for capturing word order (although see section 3), so discriminating between e.g. *he has left* and *has he left* may be problematic.
9. Polymorphic exceptions. E.g. *that* is both a DEM and C, so a possible polymorphism; there is a partial overlap between AUX and Yr, and a possibility of *have* being AUX/IR (for a systematic treatment of polymorphism, see section 11.4).
10. Specifying Y's in the formulas to its subtypes wherever possible.
11. Are there more S schemas besides those in (51)?

## 11. Type theory and UG

### 11.1. Introduction

Obviously, Tab. 1 can be somehow interpreted in a type-theoretical context. We will start by positing universes **Rel** and **Arg** (for morphosyntactic relation and argument, resp.; henceforth, all types we consider are morphosyntactic (or grammatical) unless specified otherwise). Preliminarily, since some formulas may belong to **Rel** and others to **Arg** only, it is useful to posit the maximal(ly universal) universe **U**, too. In particular, since 1st order arguments are 0th order relations, all atomic formulas inhabit **Rel** while only some inhabit **Arg**; however, because all formulas signify relations (by (13)), we have

**Arg** <: **Rel** = **U**.

Now write a formula $A(B)$ ($A$ : **Rel**, $B$ : **Arg**). Do relations depend on arguments, vice versa, both or neither? By (16), arguments depend on relations, so we need dependent types, i.e. a modern type theory; in particular, we will use Martin-Löf's type theory (MLTT). Per Martin-Löf (1984), $A(B)$ gets the preliminary form $(A,B)$. The main dependent type constructors in MLTT are $\sum$ and $\prod$. With $(A,B)$, $\prod$ would reverse the natural order by taking **Rel** for the domain and **Arg** for codomain, so we will use $\sum$ (the dependent pair type constructor) instead. Thus the formula map from L to MLTT is $A(B) \mapsto \sum(A,B)$.

Some examples. 1. Let $A$ : **Rel**, $B$ : **Arg**($A$), where **Arg**($A$) depends on $A$ by (16), then $\sum(\mathbf{Rel},\mathbf{Arg})$ is the cross-linguistically and morphosyntactically universal dependent type. 2. Let D : **Rel**, X : **Arg**. Then $\sum(D,X)$ is likely universal by (37), (38), (45). Because of selectional restrictions, it is also a dependent type: $A_{\text{COUNT}}(\text{milk})$ and $A_{\text{COUNT}}(\text{stone})$ do not have the same type (since only the latter is wf). Without selectional restrictions, $\sum(D,X)$ degenerates to the non-dependent pair type $(\sum x{:}D)X$. With selectional restrictions, specific formulas like $\text{RED}_P(X\backslash\text{PRO})$ are of the type of dependent pairs $(\sum x{:}A)B(x)$, since the formulas type(dness) will depend on $x$. For example, $\text{RED}_P(\text{spruce})$ is admissible (i.e. wf) by the selectional restriction P but $\text{RED}_P(\text{noise})$ not.

### 11.2. Axioms

We have the following axioms for L:

(77)    $\text{o}(A) : A \leq{:} A = A$
(78)    $\nvdash A : A$

(79)
$$\frac{\text{o}(A_1, ..., A_n) : B}{A_1 \leq{:} B \ \ ... \ \ A_n \leq{:} B}$$
(double line)

By (78), typing is irreflexive. The double line indicates a symmetric relation, so the converse of the (usual, i.e. single-line-separated) inference holds as well.

### 11.3. Context

We will use the type-theoretical notion of context (Martin-Löf 1984) as follows:

(80) For any wff $A_k$, there is a context $\Gamma^*$, a sequence of judgements of the form $\Gamma^* = (a_1 : A_1, a_2 :$

$A_2(a_1), \ldots, a_k : A_k(a_1, ..., a_{k-1}))$, where $A_{1 \leq i \leq k}(a_1, ..., a_{i-1})$ is a type subsuming (depending on) $a_1, ..., a_{i-1}$.

(81) Any proof (object) $a_n$ of type $A_n$ corresponds to an element of a context $\Gamma$; furthermore, there is a context $\Gamma^n$ s.t. $a_n$ corresponds to its maximal (i.e. final) element "$a_n : A_n(a_1, ..., a_{n-1})$".

A wff may be a part of several contexts. If a context consists of multiple sentences, they form a supersequence $(\Gamma_i)_{1 \leq i < m}$ in the order we encounter them. Below we show how to maintain wellorder $(\Gamma, \sqsubseteq)$ in contexts.

Examples (by (12)). 1. For THE(man), $\Gamma = (man : \text{man}, \textit{the man} : \text{THE}(\text{man}))$. 2. For "$(Y_x(\text{SMILE}))(\text{THE}(\text{man}_x))$; $(Y_x(\text{LEAVE}))(\text{he}_x)$", $\Gamma_1 = (man : \text{man}, \textit{the man} : \text{THE}(\text{man}), \textit{the man smiled} : (Y_x(\text{SMILE}))(\text{THE}(\text{man}_x)))$, $\Gamma_2 = (he : \text{he}_x, \textit{he left} : (Y_x(\text{LEAVE}))(\text{he}_x))$, and supersequence $\Gamma = (man : \text{man}, \textit{the man} : \text{THE}(\text{man}), \textit{the man smiled} : (Y_x(\text{SMILE}))(\text{THE}(\text{man}_x)))$, "$\textit{the man smiled}; \textit{he left}$" : "$(Y_x(\text{SMILE}))(\text{THE}(\text{man}_x)); (he : \text{he}_x, \textit{he left} : (Y_x(\text{LEAVE}))(\text{he}_x))$"). Since $\textit{he left}$ is a sentence, it has a separate context $\Gamma_2$, which must be built up before we can build $\Gamma$; this is indicated by parentheses around "$he : \text{he}_x, \textit{he left} : (Y_x(\text{LEAVE}))(\text{he}_x)$" in $\Gamma$. While (18) ensures a wellorder $(\Gamma, \sqsubseteq)$ for sentences, the notation does the same for supersequences.

L, its interpretation, and its relation to NL provide us with the following principles:

(82) Formulae-as-types interpretation (Curry-Howard isomorphism) [by (11) or (12)],
(83) For each wff, there is a proof of its well-formedness [by (12), (80)],
(84) All wffs are inhabited in some NLs [by (83), (80)].
(85) All subformulas of a wff are wf [by subformulas $\subseteq$ formulas = wffs].

Not all formulas of L are universal; this is equivalent to saying that some of them will be uninhabited in some languages.

### 11.4. Polymorphism

Ambiguity and underspecification are pervasive features of languages. In L, we handle them with type polymorphism, of which in NL there are 2 main kinds:

– Term underspecification ("data type polymorphism" in computer science).
– Argument ambiguity (e.g. "subtype polymorphism" in computer science).

We consider argument ambiguity first. Notice that

(86) a judgement $o(A_1, ..., A_n) : C$ is a polymorphic binary relation of type $(X \in \{A_1, ..., A_n\}, C)$.

Martin-Löf (1984) handles polymorphism with universes[3], so in MLTT one would fix a universe $U = \{A_1, ..., A_n\}$ and the relation's type would be written $(\sum_{X \in U} C)$ or $\sum(U, C)$. By (79) and (86), a number of rules in Tab. 1 (and possibly elsewhere) resort to subtype polymorphism.

Term underspecification is resolved by what we call /-types. Most /-types we analyze are **flexemes**. Lexeme is a lexical type; flexeme (or "flexible" – Luuk 2010) is a lexical type polymorphic over at least two other lexical types between which no subtyping relation holds[4]. An underspecified term $b$

---
3   Calculus of Constructions (Coquand and Huet 1988) offers a more explicit support for polymorphism.
4   As morphemes are atomic signs (form-meaning correspondences), we assume them to be lexemes (have a lexical will be typed with a /-separated string of symbols of all types over which *b*'s type is polymorphic. We will always write " $A_1/.../A_n/*$ " for "a type polymorphic over $\{A_1, ..., A_n\}$ and possibly some other types", so $A_1/.../A_n/ = *A_1/.../A_n/ \neq *A_1/.../A_n/* \neq A_1/.../A_n/*$ (by (7) and Tab. 2). For example, the following typing judgements are correct: *love, run* : X/R/*; *love, run* : X/R[/]; *love, run* : R/*; while these are incorrect: *love, run* : N/V*; *love, run* : X/R/A*; *love, run* : X/ (X/R*, though technically not incorrect, is ambiguous between R and RS). The linguistic-typological importance of flexemes lies in the fact that several Wakashan, Salishan, Munda and Malayo-Polynesian languages are claimed to lack e.g. A, ADV, N and/or V and have lexical ("flexical") categories X/*, R/* and possibly other similar ones instead (see Luuk 2010 for a survey). Cf. Czaykowska-Higgins and Kinkade (1998:36, on Salishan): "(1) all full words, including names, may serve as predicates and may be inflected using person markers /–/, and (2) any lexical item can become a referring expression by positioning a determiner in front of it". Flexemes are more widespread than we think even in better-known languages (of which there are ca 200 of 7000 – Nordhoff 2012). For example, the following

– *especially nuts are very good*
– *nuts are especially good*
– *very nuts are especially good,*

suggest a ADV/(Q|D) polymorphism for *especially* (and *particularly*), which (unlike normal adverbs like *very*) seem to conform with (43) as well as (Q|D)(X\PRO).

We continue by giving rules for /-types. In each rule, all letters except $k, m$ denote constants.

/-introduction:

$$\frac{\Gamma, \Gamma_1 \vdash a : A_1 \quad \ldots \quad \Gamma, \Gamma_n \vdash a : A_n \quad A_{1 \leq k \leq n} \not\leq: A_{1 \leq m \leq n} \vee m = k \quad n \geq 2}{A_1/.../A_n/* \text{ type} \qquad \Gamma \vdash a : A_1/.../A_n/*}$$

A /-type is defined wrt. at least two other types $A_1, \ldots, A_n$ between which no subtyping holds ($\not\leq:$ ≔ "is not a subtype of"; since empty /-types are not allowed, a single rule suffices for /-formation and -introduction).

/-elimination:

$$\frac{\neg wf(B*A_{2 \leq k \leq n}*\backslash(/*)) \quad \Gamma \vdash a : A_1/.../A_n/* \quad n \geq 2}{\Gamma, b : B*A_1*\backslash(/*) \vdash a : A_1}$$

The active (or specified) type of a term *a* of type $A_1/*$ is determined by its context (i.e. in the context of $b : B*A_1*\backslash(/*)$, *a* is typed $A_1$ under the premisses).

---

representation, map to lexicon). However, not all atomic signs (and hence, lexemes) are morphemes if we assume idioms to be atomic. Atomic signs cannot be decomposed into more elementary form-meaning correspondences, but atomic signs' form or meaning may be complex (idioms and case markers are examples, resp. – the complexity is evident if we consider the formulas of idioms or the arguments of cases in our representation).

## 12. Semantics

Compositional semantics (i.e. sentential, phrasal and morpho-semantics) is built into L by virtue of

(87) morphosyntactic types as categories of semantic values (by (1)),
(88) a mapping from morphemes and words[5] to morphosyntactic types,
(89) implementing Frege's principle of compositionality ((15), (18)-(19), (21), and complex types),
(90) selectional restrictions (section 7),
(91) coreference (section 5).

As semantics is built into L, we do not need a separate representation for it. However, if a specific (e.g. a truth-functional) one were desired, it could be (presumably depending on the representation) composed with L. As said, there is no need for it (and thus also no need for a "syntax-semantics interface" as far as compositional semantics is concerned). More generally, as witnessed by (87)-(91), morphosyntax and compositional semantics are two sides of the same coin, so an integrated treatment should be preferred to the artificial division of linguistic labor.

## 13. Universality

Obviously, one can model any NL grammar with such methods. What about UG? In L, universality is accounted for by

1. The key idea: **the universality of nonempty subsets** of certain sets (supertypes) of individual linguistic categories (types) such as N, V, A, etc. To this end, universal supertypes X, R, etc., have been employed. This principle (combined with polymorphism – see below) accounts for the universality of several atomic formulas in Tab. 1.
2. Polymorphism (e.g. for X, R and subtype definition).
3. Metaoperators |, *, ., etc. (e.g. the universality of $A|B$ entails neither the universality of $A$ nor of $B$). The operators are the main contributors to the universality of complex formulas in Tab. 1.
4. A semantically integrated treatment of morphosyntax (instead of morphology and syntax separately). Importantly, UG may have substantive morphosyntactic universals without universal morphological or syntactic features (cf. the universality of CAS|ADP).
5. The possibility to drop outermost relation symbols without affecting the well-formedness of the remaining subformula (by (85) – that the remaining string is a subformula is a plausible conjecture (see fn. 2)).

## 14. Related work

This work should fit into the existing tradition of applying modern type theories on NL (e.g. Asher 2014; Luo 2010; Ranta 1994), while being sufficiently distinct from it. The main difference from Luo and Asher is that they analyze only semantics, while major differences from Ranta are the focus on UG and the representation I develop (Ranta 1994 is a rather direct application of MLTT on NL). In addition, the very notion of morphosyntactic type on which L is based seems quite unique. The main (both philosophical and methodological) difference between Ranta (1994) and the present approach is that the former is more an application of MLTT on NL, while the latter is a description of NL in dependent and polymorphic type theory via L (with MLTT chosen as the well-known and comprehensive theory I am most familiar with). Ranta (2006) is skeptical of the idea of UG that the present paper defends. This work is probably unique in bridging type theory and linguistic typology.

---

5    Morpheme ≔ smallest meaningful NL unit; elementary word ≔ smallest meaningful unit understood without context (Luuk 2010).

## 15. Conclusion

We have described UG (and more generally, NL grammar) with dependent and polymorphic types. This is very different from the traditional Chomskyan approximation of UG with rewrite rules and/or syntax trees. The main differences from existing type-theoretical approaches to NL (e.g. Lambek 1958; Montague 2002; Asher 2014; Luo 2010; Ranta 1994) are a focus on UG, an integrated treatment of morphosyntax and compositional semantics, the formalism (or logic) L, and an employment of polymorphic and dependent types. Beyond pure description, the parsimonious specification of NL offers new perspectives for cognitively plausible (i.e. possibly (or even plausibly) human) NL processing and understanding algorithms.

**Acknowledgements.** I thank Erik Palmgren for inviting me to Stockholm University's Logic Group and Roussanka Loukanova for her feedback in a seminar there in December 2014. The work was supported by IUT20-56 "Computational models for Estonian", the Swedish Institute, and the European Regional Development Fund through the Estonian Center of Excellence in Computer Science, EXCS.